# Problems in AI, their roots in philosophy, and implications for science and society

M.J. Velthoven[1,2], E.J. Marcus[3,4]

[1] EY, International Tax and Transaction Services, Amsterdam, the Netherlands.
[2] Amsterdam Law School, Section Tax Law University of Amsterdam.
[3] AI for Oncology, Netherlands Cancer Institute, Amsterdam, the Netherlands.
[4] Informatics Institute, University of Amsterdam, Amsterdam, The Netherlands.
Correspondence to: max.velthoven@nl.ey.com, e.marcus@nki.nl

**Abstract**
Artificial Intelligence (AI) is one of today's most relevant emergent technologies. In view thereof, this paper proposes that more attention should be paid to the philosophical aspects of AI technology and its use. It is argued that this deficit is generally combined with philosophical misconceptions about the growth of knowledge. To identify these misconceptions, reference is made to the ideas of the philosopher of science Karl Popper and the physicist David Deutsch. The works of both thinkers aim against mistaken theories of knowledge, such as *inductivism*, *empiricism*, and *instrumentalism*. This paper shows that these theories bear similarities to how current AI technology operates. It also shows that these theories are very much alive in the (public) discourse on AI, often called *Bayesianism*. In line with Popper and Deutsch, it is proposed that all these theories are based on mistaken philosophies of knowledge. This includes an analysis of the implications of these mistaken philosophies for the use of AI in science and society, including some of the likely problem situations that will arise. This paper finally provides a realistic outlook on Artificial General Intelligence (AGI) and three propositions on A(G)I and philosophy (i.e., epistemology).

**Statements and Declarations**
Competing Interests: The authors declare no competing interests.

**Acknowledgements:**
We thank Samuel Kuypers, Charlotte Siemons, and Yoni Schirris for their valuable comments on earlier drafts of this paper.

# Introduction

Artificial Intelligence (AI) is one of the hottest topics of the moment.[1] Specialized applications of AI continue to expand and improve, making AI increasingly relevant as an emergent technology. Forms of AI are already being used across science and society (Hadwick, 2022). Whilst the fruits of AI are warmly received, there is a lack of attention for the philosophical aspects of AI technology and the use thereof. This deficit is generally combined with philosophical misconceptions about the growth of knowledge. Unfortunately, these are not merely theoretical problems. In this paper, we outline why these deficits are in fact real problems with implications for science and society.

This paper refers to the ideas of the philosopher of science Karl Popper and the physicist David Deutsch (section 1). Part of their work is aimed against mistaken theories of knowledge such as *inductivism*, *empiricism*, and *instrumentalism*. This paper shows that these theories bear similarities to how current AI technology operates. It also shows that these theories are very much alive in the (public) discourse on AI, often by the name of *Bayesianism* (sections 2 and 3). With reference to Popper and Deutsch, this paper argues that all these theories are based on mistaken philosophies of knowledge. Identifying such deficits is an important task of philosophy: to point out what we do not know, why some of our existing knowledge is mistaken, and what problems our ignorance gives rise to. This paper analyses the implications of these mistaken philosophies for the use of AI in science and society (section 4).

Despite the philosophical problems associated with AI, we recognize the potential of AI to bring great benefits in many fields. This potential makes it even more important to have sufficient awareness of how current AI works and what it can and cannot do. In this paper, we outline some of the likely problem situations that will arise as a result of the lack of such awareness. This paper also provides an outlook to the future, including a realistic outlook on Artificial General Intelligence (AGI) (section 5). This part of the paper proposes that decent public policy on AGI requires a sound underlying philosophy and that progress in philosophy is a key requirement for AGI. The paper ends with a conclusion and three propositions (section 6).

# 1   A brief introduction to the ideas of Karl Popper and David Deutsch

*''I think (like you, by the way) that theory cannot be fabricated out of the results of observation, but that it can only be invented.''* – Albert Einstein in a letter to Karl Popper.[2]

This section introduces the key aspects of the ideas of Karl Popper that are most relevant to this paper. Karl Popper (1902 – 1994) was a philosopher and academic who is mainly known for his work on the philosophy of science (in particular his Falsification Principle) (Magee, 1973; Popper, n.d., 1962). This introduction should start with two upfront warnings. The first warning is slightly obvious: Popper's ideas have immense depth and reach. One commentator (Magee, 1973) described them as *''interconnecting parts of a single explanatory framework which extends to the whole of human experience''* (p. 16). As a consequence, any attempt at summarizing these ideas is like trying to fit an entire year's grape harvest into one glass of wine. This is why reference is made to others for a more detailed discussion of Popper's works (Magee, 1973; Popper, 1976).

---

[1] The title of this paper is a reference to Karl Popper's paper 'The Nature of Philosophical Problems and Their Roots in Science', which is included in his Conjectures and Refutations: The Growth of Scientific Knowledge (footnote below). An excellent discussion of this paper can be found in Brett Hall's podcast TokCast, episode 19 (Mr. Popper's Problems).

[2] Albert Einstein in letter to Karl Popper, see Appendix to (Popper, 1934), p. 482. The full quote is: ''Altogether I really do not at all like the now fashionable [modische] 'positivistic' tendency of clinging to what is observable. I regard it as trivial that one cannot, in the range of atomic magnitudes, make predictions with any desired degree of precision, and I think (like you, by the way) that theory cannot be fabricated out of the results of observation, but that it can only be invented.'', see: (Popper, 1934).

The second warning is even more important. Despite his intellectual achievements, Popper's ideas have generally been remarkably *unsuccessful* in academia as well as in the rest of the world. On the contrary: the ideas of thinkers whom Popper opposed have been much more successful and continue to be so up and until this day. In the area of the growth of knowledge, one should particularly think of Ludwig Wittgenstein, Thomas Kuhn, Francis Bacon, and various postmodernist thinkers.[3] As a consequence of the success of this opposition, Popper's ideas stand diametrically opposed to the philosophical ideas that govern the science of today. Adding an additional complication is the fact that many of Popper's ideas may seem obvious or even trivial upon first encounter. However, it should be kept in mind that current mainstream ideas on science and its methodology are completely different from Popper's philosophy.[4] Hence, Popper's philosophy is anything but trivial or obvious. If the reader has never stumbled upon the ideas of Popper and Deutsch before, this paper is best read under the assumption that their ideas stand in full opposition to the reader's prior ideas on the methodology and philosophy of science.

So, what are these 'radical' ideas? One of Popper's former students, the Israeli philosopher Joseph Agassi, found Popper's biggest achievement to be that *''Popper took Socrates' motto of 'I know that I know nothing' and brought it to modern science''*.[5] This idea can be found in one of the more famous aspects of Popper's work, namely his attack on the scientific method of *induction*. In short, induction is the idea that universal statements can be derived from specific observations. Most people who have heard of Popper only know him for one specific element of his attack on induction. This is Popper's proposition that *no matter how many white swans we see, seeing more of them never proves that all swans are white*.[6]

Popper combined this attack on induction with his counterproposal for an alternative method called *falsification*. This idea holds that universal statements can never be proven *right* by specific observations. Universal statements can, however, be proven *wrong* by specific observations: if we see even one black swan, this means that *not all swans are white.*[7] Thus, in the spirit of Socrates, whilst we do not know anything to be *true*, we only know something to *not be true*. And to make matters worse, Popper was of the view that even this limited knowledge was always tentative and error-prone. Popper drew another important conclusion: not only do singular observations not increase the truth content of a statement, *they also do not increase the probability of a statement being true*. The fact that we see yet another white swan does not increase the probability of the statement that all swans are white.[8] This is a crucial proposition for this paper. In subsequent sections we will see that much of current AI technology works by assigning additional probability to a statement if it has been found to be in line with an observation (or, vice versa, decreasing the probability if a contradicting observation is found). AI technology therefore operates in line with the very same ideas on probability and truth which Popper rejected.

---

[3] Popper's books mentioned in the references contain attacks on all of these thinkers (i.e., their ideas).

[4] As an example: in chapter 4 of (Popper, 1994), Popper warned against the dangers of too much specialization in science. Popper provides a dystopian view of what science would become if scientists would specialize themselves too much within one area. In the authors' view, much of today's scientific world resembles the future which Popper warned against.

[5] Agassi made this remark in a conversation with one of the authors of this paper. We also refer to (Benesch, 2012), who referred to Popper as the Viennese Socrates.

[6] For the origins of the swan metaphor and how it was created by the writer Nassim Nicholas Taleb (Taleb, 2009), see: https://en.wikipedia.org/wiki/Black_swan_theory.

[7] Concerning our earlier proposition on the complex nature of Popper's work, we should note that this part, in particular, is a highly simplified depiction of Popper's ideas. Readers interested in a more detailed elaboration are referred to Magee, who writes: ''So although Popper is what might be called a naïve falsificationist at the level of logic, he is a highly critical falsificationist at the level of methodology. Much misunderstanding of his work has sprung from a failure to appreciate this distinction.'' See (Magee, 1973), chapter 2: Scientific Method – the Traditional View and Popper's View.

Popper proposed that knowledge grows exactly the other way around than is commonly thought in science. Observation does not precede theory: theory precedes observation. In fact, Popper famously thought that *all observations are theory-laden* (Popper, 1962), chapter 1; (Popper, 1934), chapter 3. If not for observation, what should we revert to as the basis of our theories? For Popper, the source of new theories could be anything: a guess, a hunch, or even a creative impulse. Regardless of how we come up with them, our made-up theories (also: *conjectures* or *explanations*) are subsequently tested against the real world, whereby theories can only be (tentatively) *falsified*, never confirmed, proven or shown to be 'probable'. In addition to these tests, theories are also subject to criticism by ourselves and others, leading to a continuous cycle of tentative explanations (*conjectures*) and tests/criticisms (*refutations*). According to Popper, this is how knowledge grows both in an individual mind as well as in society generally.[9] This is also why Popper refuted the philosophy of *instrumentalism*. Instrumentalism broadly speaking is the idea that knowledge is a tool: as long as knowledge does the job, it works and it does not have to be questioned or explained.[10] There are at least two major issues with instrumentalism from the perspective of Popper's philosophy. Firstly, instrumentalism does not lend itself to falsification: a piece of knowledge either works or it does not but is never falsified. Secondly, instrumentalism is *explanationless* science, whereas Popperian science is all about explanations.[11] In instrumentalism there is no explanation, only a result indicating whether or not a theory's prediction was correct.[12]

A real-world example of instrumentalism in today's science can be found in quantum physics. Physicists engaged in quantum theory often adhere to the so-called "Shut up and calculate" approach. As the name suggests, this approach dodges any questions or genuine explanations on how the quantum world works (the 'shut up' part). Instead, the approach merely provides the expected results of quantum experiments (the 'calculate' part). Similarly, current AI technology also only engages in the 'calculate' part, without engaging in any deeper questions or explanations. This means that AI by itself cannot advance science as it cannot explain why certain theories should be preferred over others. AI can nevertheless provide an important supportive role in science, just like calculators which aid in scientists in making computations. However, AI itself does not formulate new scientific explanations.

---

[8] Podcaster Brett Hall provided a wonderful argument for this proposition in an episode of his *TokCast*. Hall proposes that the ''contrapositive'' of the statement ''all swans are white'' must be that ''all non-white object are not swans''. If the observation of a white swan increases the probability of the statement that ''all swans are white'', this means that the observation of a *red pen* (a ''non-white object'') increases the probability of the statement that ''all non-white objects are not swans'' and hence also of the statement that ''all swans are white''. See TokCast, Episode 187, Red Pens and Fallibilism.

[9] Knowledge, in Popper's and Deutsch's view, can be explained in several, but equal, ways. For purposes of this paper, the most pragmatic description is that knowledge consists of explanations that best solve problems at hand. From this definition it follows that solving (scientific) problems causes knowledge to grow. For more elaborate discussions on knowledge and how it grows, see (Marletto, 2021), p. 34-37; (Popper, 2013).

[10] Popper writes: ''Since 'right' here means 'applicable', this assertion merely amounts to saying 'Classical mechanics is applicable where its concepts can be applied' - which is not saying much. But be this as it may, the point is that by neglecting falsification, and stressing application, instrumentalism proves to be as obscurantist a philosophy as essentialism. For it is only in searching for refutations that science can hope to learn and to advance. It is only in considering how its various theories stand up to tests that it can distinguish between better and worse theories and so find a criterion of progress.'', see: (Popper, 1962), chapter 3: Three views concerning human knowledge, paragraph 5: Criticism of the instrumentalist view.

[11] Popper generally refers to *conjectures,* whereas Deutsch refers to *explanations*.

[12] For a discussion why explanationless science cannot advance science, see (Deutsch, 1997).

This paper is not the first endeavor to bring Popper's thinking into the area of AI. The physicist David Deutsch has already done so on several occasions, for example in Ch. 7 of his book (Deutsch, 2011) as well as in later essays (Deutsch, 2012a, 2012b). Deutsch is widely regarded as the father of the quantum computing, for which he has been awarded several prizes, including the *Breakthrough Prize in Fundamental Physics*. In his works, Deutsch consistently comes up against the same philosophical ideas which Popper objected to (positivism, inductivism, etc.). Deutsch also blames deficiencies in philosophy as the explanation for the lack of progress in artificial creativity, which is a point particularly relevant for this paper. Deutsch recognizes the huge potential of specialized applications of AI and is of the view that AGI should be achievable. However, in Deutsch' view, achieving AGI would firstly require a breakthrough in philosophy which would answer some important Popperian questions. For example: *how* do humans formulate explanations? And how do we go about criticizing them? There may very well be progress in AI before these questions are answered but according to Deutsch we should not expect this progress to result in something resembling human creativity or AGI. As Deutsch wrote: *''Expecting to create an AGI without first understanding how it works is like expecting skyscrapers to fly if we build them tall enough.''*(Deutsch, 2012a)

## 2   How does current AI technology work?

This section provides a short overview of the functioning of current AI technology. Although different variants exist, a simplified depiction of the working of many AI models would be as follows[13]:

1. Data collection and cleaning of the dataset.
2. Training on a subset of the total dataset.
3. Testing the model on another subset of the total dataset.

AI engineers and scientists often add an additional evaluation step aimed at optimizing the model architecture and other parameters. This additional step takes place before the testing step (step 3) and uses another subset of the total dataset. The heavy reliance on (sub)sets of data means that current AI is generally considered to be *data hungry*: even small models may require thousands of data points to achieve reasonable performance.

As an illustrative example, suppose that an AI model would be designed to categorize images as either a "cat" or a "dog". It is also assumed that this AI model is a classical neural network in the form of a multilayer perceptron (MLP), although the specific form is not relevant for the purposes of this paper.

1. In the first phase, a database is created which includes (thousands of) pictures of cats and dogs along with labels specifying to which class each of the pictures belongs. This database is cleaned, thereby eliminating unclear pictures or pictures that do not fall within one of the categories (such as pictures of elephants).
2. In the second phase, the model is shown batches of images along with labels specifying to which class the images belong. Subsequently, the model provides predictions of whether a certain image depicts a cat or a dog. These predictions are judged using a loss function that penalizes the model for wrong predictions. In sports terms: the model concedes a goal for every wrong prediction. The model will subsequently use its own predictions and labels to update its parameters such that it will perform better in the next iteration. Usually, the model is shown the entire training dataset several times before it has optimized its parameters to the training dataset.

---

[13] For an elaborate introduction to deep learning methods see (Goodfellow et al., 2016). For machine learning (not deep learning) see (Hastie et al., 2013).

3. The model is subsequently tested on another subset of pictures of cats and dogs which is different from the training set.

Whilst going through the above sequence, the neural network will become better at estimating classes in its training data as this is what the model is being trained to do. But how well will the model perform on the test dataset? And more importantly, will the model become better at data outside of the original dataset? In the example of a trained cat/dog AI model, the general expectation is that the model could work well on other datasets, provided the input images are sufficiently similar to what the model has seen before. In other words: if the new pictures of cats and dogs that are shown to the model are sufficiently similar to the pictures it was shown before. The similarity between a new datapoints and the training dataset is usually described as 'in distribution' or, conversely, as 'out of distribution.' AI models are expected to work well whenever the new data points fall 'in distribution' whereas this is generally no longer the case when the data point falls 'out of distribution'.[14]

The AI model's failure to apply the ''lessons'' from the dataset to entirely new situations provides the first parallel with philosophy. This failure resembles the famous *chicken paradox* put forward by the philosopher Bertrand Russell as part of his analysis of the philosophical concept of induction (Russell, 2001). Russell proposed that a chicken, accustomed to being fed daily by a farmer (the training data), will falsely infer that seeing the farmer always means that she will be fed (an 'in distribution' event). However, the day will come when the farmer does not come to feed the chicken but to feed himself, hence slaughtering the chicken (an 'out of distribution' event). For Russell, this example showed that there were serious defects in the concept of inductive reasoning.

Another class of AI models that quickly grew in popularity consists of the so-called *generative AI models*. Examples of these models include Large Language Models (LLM), such as ChatGPT, and AI art models such as DALL-E 3.[15] These models are trained slightly differently than the models we discussed before. As mentioned before, an in-depth discussion of the various inner workings of generative models is out of scope of this paper. We will therefore suffice with some general remarks instead. Whilst there is a huge variety in generative AI models, they all have one trait in common: they attempt to create samples that look like the distribution they are being trained on. For the LLM, this means that certain words (or 'tokens') are masked out in the input data. For example, AI trainers may come up with a phrase such as "The [x] ate the dog food." and ask the model to produce the correct word for [x] in view of that context (in this case, [x] would stand for ''dog''). For the art models, the dataset consists of a large collection of images. Subsequently, AI trainers may add 'noise' to the images and ask the model to reproduce the original, clean image. By performing this task thousands or even millions of times, the model 'learns' the distribution of pixels of the training dataset.

## 3  An analysis of current AI technology in view of the works of Popper and Deutsch

The previous section already provided the first hints of the similarities between the working of AI models and the philosophies which were rejected by Popper and Deutsch. In particular, the functioning of current AI much resembles the philosophy of *inductivism* (or its modern instantiation, *Bayesianism*, which we will analyze below). As Popper has shown, the method of inductivism cannot hope to create true new 'universal' or even 'more probable' statements based on individual instances. Yet this is precisely how AI systems work. Paired with the wrong philosophies of knowledge, this gives many people the wrong impression, namely that what AI does is similar or even identical to how humans generate new knowledge.

---

[14] For a recent survey see (Liu et al., 2023). Furthermore, the related website provides an overview of recent work on out of distribution detection: https://out-of-distribution-generalization.com/

[15] For a clear overview of various algorithms used for language modeling, see (Phuong & Hutter, 2022).

For instance, let us again look at the cat/dog classifier model, where we now assume that the dogs in the dataset are almost always pictured *sitting outside on grass fields*. Whilst this is likely not 'intended' (by the AI nor its human supervisor), the AI model can inadvertently be 'instructed' to assign a higher probability to there being a dog in the picture if there is a green background. In more extreme cases, the model may even render it fully certain that there is a dog in the picture if there is a green background. Even though the model itself does not develop an explicit theory, the model might thereby effectively operate in line with the universal statement *that every picture with a green background is a picture of a dog.* If so, the model has done something that is identical to the method of inductivism: the model has formulated a general statement ('every picture with a green background is a picture of a dog') based on specific observations (ten thousand pictures of dogs on grass fields). In Popper's jargon: the model may conclude that all swans are white, just because it has seen sufficient pictures of white swans. We should emphasize that the model strictly speaking does not 'conclude' anything or 'formulate' a general statement. All the model effectively does is develop a decision tree based on which a sufficient number of green pixels results in the output that input picture is a picture of a dog. The model does this without formulating any underlying explanation.[16] For an in-depth discussion of how humans can still find explanations of what AI models do, see (Marcus & Teuwen, 2024).

Some people might argue that current AI does not resemble the philosophical theory of *inductivism* but instead resembles a theory often referred to as *Bayesianism*. However, we would argue that these theories are ultimately two sides of the same coin. The reason for this is that the Bayesian approach in AI is ultimately also based on the inductivist method. In Bayesianism as applied in AI, every individual instance results in an adjustment of the probability distribution. This means that specific instances impact the predictions for future events. The AI model's performance generally depends on the data we provide it with. We previously illustrated the workings of AI with reference to a cat/dog classifier. However, the importance of our philosophical observation is more obvious when the Bayesianism functioning of AI is illustrated in another field, namely healthcare. An example of a faulty diagnosis in this field could be a perfectly healthy patient who is diagnosed with a disease which the patient does not have. There can be various reasons for such a faulty diagnosis. For example, suppose that a dataset has been created based on a big number of diagnoses of patients that potentially have a certain disease. Unfortunately, such datasets frequently contain severe imbalances. The dataset may contain a non-representative ratio of positive and negative cases, or a demographic build-up that is not representative for the actual target group. Imbalances in the dataset may even be found in more technical details: the academic hospitals which often provide the data for medical AI will generally have the most advanced MRI scanners available. This means that all the positive cases in which the disease is found will be on patients that have been scanned with the most MRI advanced scanner. Based on Bayenisian reasoning, the model may subsequently allocate an increased probability for finding the disease if a patient has been scanned by an advanced MRI scanner. In doing so, the model is effectively operating in line with the universal statement that *that every patient that has been scanned by an advanced MRI scanner has an increased probability of the disease.* This shows that Bayesianism, as also applied in current AI models, is prone to similar philosophical deficits as inductivism.

---

[16] More precisely, neural networks make large cascades of if-then statements. In these statements, it may combine many (or even all) of the previous layer's outputs. The resulting decision trees can become complex enough to correctly approximate 'dogs' in many circumstances. However, in such cases, there is still no underlying explanation, and dogs may fail to be detected in scenarios where humans easily do so. For instance, dog-detection models may fail in a famous computer vision problem of image classification on 'chihuahua versus muffin', unless they are specifically trained to be good at it. For a more in-depth discussion regarding neural networks and decision trees, see (Aytekin, 2022).

The philosophical ramifications of this extend to the use of AI in the growth of human knowledge, including science and art. From the perspective of the philosophy of Popper and Deutsch, both inductivism and Bayesianism (as applied in AI) cannot generate truly new scientific knowledge. Instead, their philosophy proposes that such knowledge can only grow by making new explanations (conjectures). The cat/dog AI model did not generate new explanations during its learning process (nor did the medical AI model). The model does not *explain* anything: it simply gives us numbers indicating the probability that a particular image represents one class or another. Any explanation of its workings is only provided afterwards (by people). In contrast, *people* learn what a dog or a cat is by virtue of explanations, not by 'extrapolating' from a collection of thousands of pictures. Therefore, AI should be considered as an instrument rather than a creator of new explanatory knowledge. Humans remain the only source of new explanations that has yet been found. Deutsch even refers to humans as 'universal explainers': entities that can generate explanations on any possible subject matter. In Deutsch' view, any AI system that would create new explanatory knowledge would be an AGI (more on this below). Current AI technology might very well help to solve a problem or provide a hint on which new explanatory knowledge could be based. However, even if the AI provides a literal explanation of some kind, it is ultimately up to the AI's human user to recognize this as an explanation. Only thereafter will the new explanatory knowledge be truly used as an explanation.

For completeness's sake we note that our proposition remains unchanged for all AI technologies which are currently known, including generative AI. As mentioned before, generative AI models optimize their training task whenever they create samples that are close to the distribution of the training dataset. This is still entirely different than the way in which humans generate new ideas. When people generate new ideas, their new ideas may be completely unlike any idea that was ever seen before. For instance, Einstein's theory of general relativity has elements completely unlike Newton's theory of gravity which preceded it. One of these differences is that Einstein's theory has no force of gravity but considers only the curvature of spacetime. Similar creation of new ideas occurs in other fields of human output, such as art. For example, Picasso's works were definitely not mere mixtures of samples from a distribution of art that came before.

Although generative AI models' training procedure forces these models to adhere to the training data, people may still use these models for the creation of new texts or images. For instance, by prompting an LLM on a new, previously unwritten, piece of text, the model can be made to produce sentences fitting for this prompt. Suppose we wish to write an entirely new story. If so, we can go through this cycle of prompting and subsequently, manually, criticizing the new text which is generated by the AI. In doing so, we utilize our (human) explanations and criticisms in the creation of the new text. In this sense, the AI models can be *instruments* in creating new (explanatory) knowledge. However, the models should still be seen as instruments (i.e., advanced typewriters or paint brushes), with people remaining at the creative steering wheel. The AI remains just a *tool*; the human the *universal explainer*.

## 4   Implications

In the previous sections we proposed that current AI technology operates similarly to mistaken philosophies of knowledge. Does this mean that we are opposed to the use of this AI technology? Quite the contrary: we recognize the many potential use cases for AI and see no convincing reason against it, provided that certain guardrails are in place. Unsurprisingly, these guardrails address the issues raised in the previous section. If these are not properly dealt with, the use of AI in combination with the aforementioned misconceptions on the growth of knowledge can result in disaster.

As a start, actions taken regarding AI should never be based on the misconception that *the operation of AI is similar to how people think*. This proposition has major implications for science and society. In the design of organizations and of policies, AI should be seen as a new technology (a tool), not as a replacement of human intelligence. Some might object and propose that AI might nevertheless perform tasks currently performed by humans and in doing so replace human jobs. We would argue that is not a new phenomenon: the typewriter and the telephone have evidently taken over work that was previously performed by humans. However, crucially, no one would dare say that these technologies have replaced human thinking. The same restraint should apply to AI.

Treating AI as ''just'' a technology also implies that the ultimate responsibility to explain the applications of AI should remain with people, not with the AI itself. This implication is relevant for all applications of AI regardless of whether AI is applied in the private or public domain. However, public policy (laws) on AI should particularly take note of these deficiencies. One important reason for this is that governments have the unique capability of creating universal laws which govern AI and its development in its entirety. This means that, while misconceptions about the growth of knowledge are widespread, incorporating these misconceptions into laws may lead to particularly grave consequences. Another reason is that the philosophical deficiencies of AI will particularly manifest themselves in areas which are generally regarded as crucial in public governance. Below we provide a few examples, with reference to the mistaken theories of knowledge called *inductivism* and *instrumentalism*.

In the prior sections we have shown that current AI technology uses *inductive* reasoning (or Bayesian reasoning, which we have shown to be identical in this regard). As shown above, induction has proven to be wrong as a philosophy of science. In a real-world application through AI, inductive reasoning can result in various problematic outcomes. The below examples show the grave consequences hereof in one area of the public policy domain, being taxation. We have chosen taxation as it is a domain in which AI is already being used extensively by governments (Hadwick, 2022).

- **Direct discrimination:** *the prior datasets used to train an AI model showed instances of tax fraud among people with nationality A. Therefore, the model assigns an increased probability of fraud for people with nationality A. As a result, people with nationality A are scrutinized more often in tax audits going forward.*
- **Indirect discrimination:** *whilst a tax audit AI model does not directly consider nationality, it implicitly does so. For example, it may operate by weighing whether a certain person has travelled to country A during the last five years.*
- **False dichotomies:** *based on prior experience, the cause of an underpayment of tax provided by an AI model is always either A or B. Therefore, its human users make the mistake of concluding that these are the only options available to explain an underpayment of tax. Upon inspection, this turns out to be wrong, as there are various other causes for the underpayment.*

As mentioned before, a closely related misconception to induction, which is omnipresent in discussions around AI, is *instrumentalism*.[17] Broadly speaking, this is the idea that anything that works (i.e., delivers results) is knowledge. In a real-world application, instrumentalism can result in various problematic outcomes. A few practical examples below, again in the public governance area of taxation.

- **Lack of explainability:** *under many legal frameworks, tax authorities are obliged to explain why a certain taxpayer has been selected for audit and why they proceed with taking certain steps (such as imposing an additional tax assessment). If tax authorities use an AI model to determine which taxpayer to audit, there is a clear risk that such explanations cannot be provided. The only explanation might be that ''the model works''.*
- **Lack of responsibility:** *there is also a risk of lack of responsibility. In the AI tax audit model, there is the risk that none of the employees of the tax authorities will take any responsibility for the outcomes of the model and will suffice by stating that ''the AI told me to do this.''*

---

[17] The primary evidence for instrumentalism follows from what is essential for publishing in AI. New articles are judged mostly (or only) based on their results on various metrics. If the model has a State Of The Art (SOTA) result (often evidenced by a bold number in a table), it is more likely to be appreciated.

How can these risks be mitigated? As a start, public policy on AI should be reflective of the fact that AI operates similar to mistaken philosophies of knowledge. In our view, it is an immense risk if AI policies are put in place in the field without such awareness. This is not a theoretical risk: there are various examples in which the use of AI in public governance has already resulted in the above issues, such as discrimination (Zaken, 2020). For example, in the Dutch *Childcare benefits scandal*, the use of AI by the Dutch tax authorities resulted in taxpayer discrimination (Hadwick, 2022). AI policies should therefore require that people remain at the *creative steering wheel* when partnering with AI. Human actors should also continue to bear the ultimate *responsibility* to *explain* the application of AI. AI technology which does not facilitate these requirements should not be used by government. These recommendations of course do not stop at the policy level: people working in the management and day-to-day application of these systems should also be sufficiently aware of this. They should be aware of their own responsibility when using AI and of the required explainability of its outcomes. It is needless to say that all of this also applies if AI is applied in other areas. For example, it should also be considered in the design of research projects if AI is used in science.

In conclusion, policies in respect of AI should consider the philosophical deficits in our understanding of current AI technology. A failure to do so could result in various adverse consequences, such as a wrong use of technology, lack of accountability and transparency, discrimination, and a mismanagement of expectations. Specifically, a misunderstanding of the underlying mechanism of knowledge creation can result in poor regulation and governance strategies.

# 5 Outlook and a realistic philosophical perspective on Artificial General Intelligence (AGI)

We end this paper with some last observations on the *holy grail* of the AI debate: the potential for AGI. Although there are different interpretations of this term, AGI generally refers to artificial intelligence which equals human intelligence. Such an AGI would be able to create new explanations for anything (a 'universal explainer', as Deutsch calls it). As will have become clear from this paper, the ideas of Popper and in particular Deutsch support the notion that AGI is possible. In fact, one of Deutsch' key propositions is that AGI *must* be possible.[18] Deutsch however concludes that AGI will not be achieved by the current progress in AI, no matter how good AI may become in specialized tasks such as modelling language or image recognition. Progress in current AI is much welcomed but should merely be seen as a further refinement based on the same technology and (mistaken) philosophy as was described in the previous sections of this article. Deutsch therefore finds that current AI technology is – if anything – moving away from AGI rather than towards it.

Unsurprisingly, we find many traces of the ''inductivist'' and positivist'' schools of thought in the current AGI debate. These traces are particularly visible in the popular idea that AI will eventually turn into AGI once AI has obtained sufficient data or computer power. Deutsch is of the view that that a breakthrough in philosophy is needed before AGI even becomes remotely possible. Such a breakthrough would have to answer questions such as 'How do people make new explanations?' and 'How do people formulate criticisms?'. Without answers to these questions, even an AI that is fed with all the computer power and data in the universe will not become an AGI.

Unfortunately, the current discourse on AGI rarely includes a clear formulation of the underlying philosophy of participants in the discussion. As a result, participants are free to make the wildest predictions about the likelihood of achieving AGI soon and its potential consequences without providing any clues on the philosophical requirements for AI to reach the level of AGI (see, e.g., (Macey-Dare, 2023)). In our view, the discourse on AGI requires a clear formulation of the underlying philosophy of its participants. Suppose that this philosophy is in line with the Popperian approach outlined in this paper. In that scenario, it should be admitted that a key ingredient of AGI is still missing. In the public discourse, this may mean that government funding for AGI projects will be hard to defend, as the public will wonder why taxpayer money is being used on projects which its own government does not believe in. The same applies to any attempts to regulate AGI, such as the AI Act (EU, 2024). As Deutsch recently remarked: *as long as we do not know how AGI will work, trying to regulate it is like trying to make a regulation for starships*.

---

[18] Deutsch bases this idea on the concept of *computational universality* (see his book (Deutsch, 2011)).

Admittedly, it is more realistic to assume that there is currently *no* underlying philosophy in the current discourse on AGI. If there is *any* underlying philosophy at all, it is likely to be in line with the ideas that Popper and Deutsch argued against. This might be the idea that ''knowledge grows by collecting objective facts'', or its AGI cousin ''AGI will be created once AI has been provided sufficient data and computer power.''. Unsurprisingly, we find both scenarios to be highly undesirable: the governance of such an important topic requires an explicit philosophy which can be criticized by others. The absence of a clear philosophy of knowledge underlying the public governance of AGI does not only run the risk of generating mistakes: it also prevents lessons being drawn from these mistakes.

# 6 Conclusion: three propositions

This paper has analyzed current AI technology and criticized its philosophical aspects. In its criticism, the paper has made specific reference to the ideas of Karl Popper and David Deutsch. It has become apparent that there are many similarities between the operation of current AI technologies and mistaken philosophies on the growth of knowledge which were criticized by Popper and Deutsch. This paper proposed that this has significant implications for the use of AI. We summarize our conclusions in the following three propositions, which should be kept in mind for any (public) policies on AI and AGI.

1. Information created by current AI is not new 'knowledge' as created by humans: humans should therefore remain at the 'creative steering wheel'. AI is an instrument and cannot by itself formulate new (scientific) theories.
2. It follows that current AI cannot by itself generate truly new explanations whereas humans can. An important implication is that human actors should continue to bear the ultimate responsibility to explain the application of AI. This will inevitably include cases in which AI is successfully used as an instrument even though its operation itself cannot be explained (black box AI).
3. Progress in current AI may result in specialized applications that bring great benefits to science and society. However, there is no reason to believe that further development of current AI technologies will result in the creation of an AI that is truly on par with human intelligence (also called AGI). In fact, the current developments in AI are actively moving away from realizing AGI.